\documentclass[conference]{IEEEtran}
\IEEEoverridecommandlockouts
\usepackage{cite}
\usepackage{amsmath,amssymb,amsfonts}
\usepackage{algorithmic}
\usepackage{graphicx}
\usepackage{textcomp}
\usepackage{xcolor}
\usepackage{booktabs}
\usepackage{url}
\usepackage{tikz}
\usepackage{subcaption}
\usepackage{siunitx}   
\usepackage{bm}
\usetikzlibrary{arrows.meta, positioning, decorations.pathreplacing, calc, backgrounds, fit, shapes.geometric}
\def\BibTeX{{\rm B\kern-.05em{\sc i\kern-.025em b}\kern-.08emT\kern-.1667em\lower.7ex\hbox{E}\kern-.125emX}}

\newcommand{\regsym}{\textsuperscript{\tiny\textregistered}}
\newcommand{\orcidlink}[1]{}

\begin{document}

\title{Dual Pose-Graph Semantic Localization for Vision-Based Autonomous Drone Racing}

\author{
\IEEEauthorblockN{
David Perez-Saura\textsuperscript{*}\orcidlink{0000-0003-2571-3165},
Miguel Fernandez-Cortizas\textsuperscript{*\dag}\orcidlink{0000-0002-3822-075X},
Alvaro J. Gaona\textsuperscript{*}\orcidlink{0009-0003-4967-4444},
Pascual Campoy\textsuperscript{*}\orcidlink{0000-0002-9894-2009}        
}
\IEEEauthorblockA{\textsuperscript{*}\textit{Computer Vision \& Aerial
Robotics Group, Universidad Polit\'ecnica de Madrid,} Madrid, Spain}\IEEEauthorblockA{\textsuperscript{\dag}\textit{Automation and Robotics    
Group (ARG-SnT), University of Luxembourg -- SnT,} Luxembourg}               

}

\maketitle

\begin{abstract}  
Autonomous drone racing demands robust real-time localization under extreme conditions: high-speed flight, aggressive maneuvers, and payload-constrained platforms that often rely on a single camera for perception. Existing visual SLAM systems, while effective in general scenarios, struggle with motion blur and feature instability inherent to racing dynamics, and do not exploit the structured nature of racing environments. In this work, we present a dual pose-graph architecture that fuses odometry with semantic detections for robust localization. A temporary graph accumulates multiple gate observations between keyframes and optimizes them into a single refined constraint per landmark, which is then promoted to a persistent main graph. This design preserves the information richness of frequent detections while preventing graph growth from degrading real-time performance. 
The system is designed to be sensor-agnostic, 
although in this work we validate it using monocular visual-inertial odometry and visual gate detections. Experimental evaluation on the TII-RATM dataset shows a \SIrange{56}{74}{\percent} reduction in ATE compared to standalone VIO, while an ablation study confirms that the dual-graph architecture achieves \SIrange{10}{12}{\percent} higher accuracy than a single-graph baseline at identical computational cost. Deployment in the A2RL competition demonstrated that the system performs real-time onboard localization during flight, reducing the drift of the odometry baseline by up to \SI{4.2}{\meter} per lap.
\end{abstract}



\section{Introduction}

\begin{figure}[t]
    \centering
    \includegraphics[width=1.0\columnwidth]{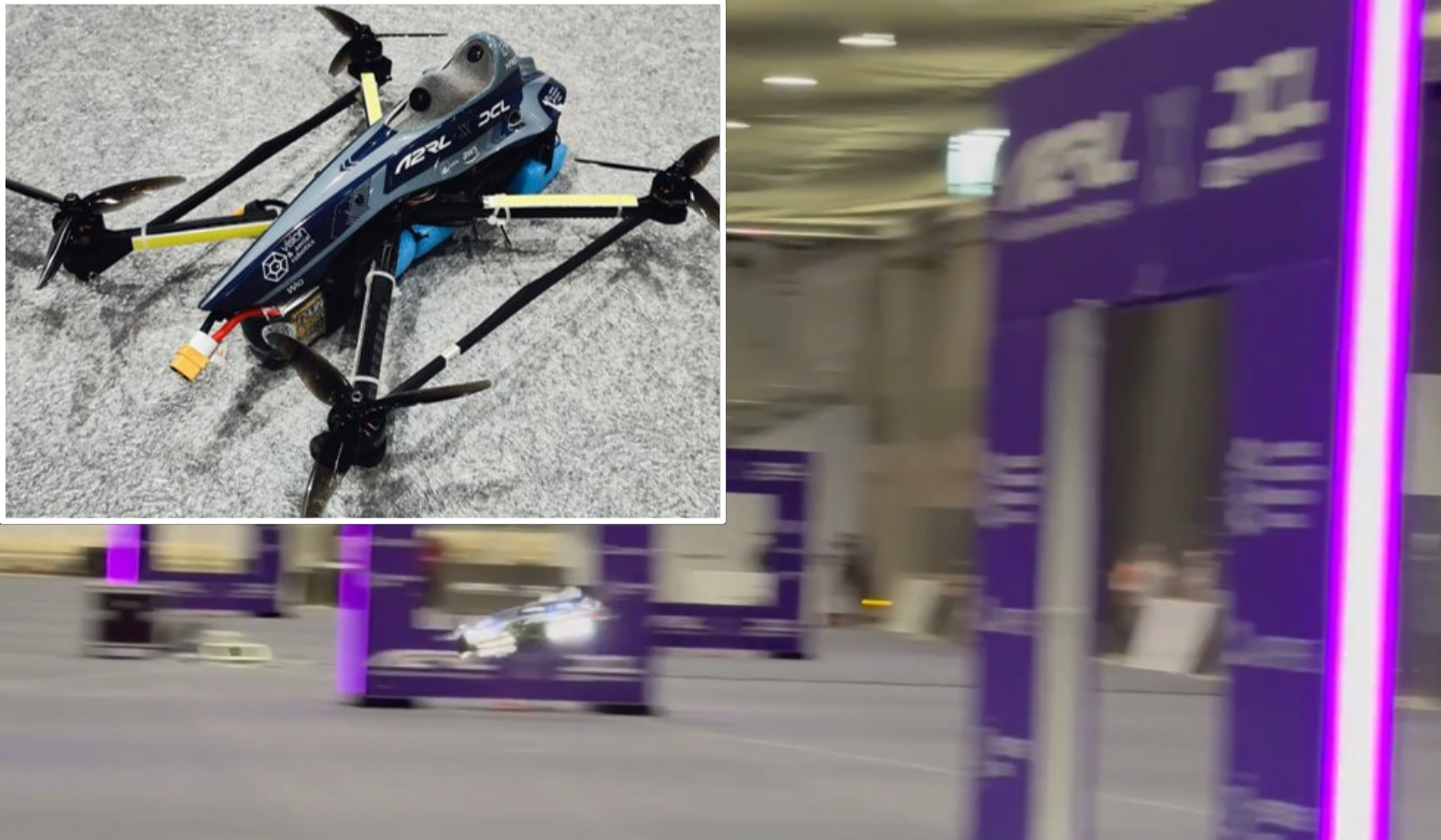}
    \caption{Drone flight during A2RL competition. The racing drone is equipped with a monocular camera for visual-inertial odometry and gate detection.}
    \label{fig:a2rl}
    \vspace{-5pt}
\end{figure}

Autonomous drone racing has emerged as a challenging benchmark for agile perception and control, pushing the limits of onboard sensing and computation~\cite{kaufmann2023champion, foehn2022alphapilot}. Competitions such as the Abu Dhabi Autonomous Racing League (A2RL) require drones to navigate through sequences of racing gates at high speed with minimal sensor payloads, often limited to a single monocular camera (Fig. \ref{fig:a2rl}). Under these conditions, accurate and robust localization is critical for trajectory planning and gate traversal, yet it remains an open challenge.

At racing speeds, visual odometry (VO) accumulates drift rapidly due to motion blur, limited feature persistence, and aggressive rotational dynamics. Feature-based SLAM systems, such as ORB-SLAM3~\cite{campos2021orbslam3} can mitigate drift through loop closure, but they rely on stable visual features that degrade in fast motion. Visual-inertial odometry approaches like VINS-Mono~\cite{qin2018vins} offer improved robustness through IMU fusion, but IMU data may not always be available or reliable in the extreme dynamics encountered in racing. Direct methods are sensitive to photometric changes caused by varying lighting on the track. None of these systems exploits the structured and repetitive nature of racing environments.

Racing tracks, however, provide a strong semantic prior: gates are distinctive, repeated landmarks whose positions define the track layout. Incorporating gate detections into the localization pipeline can provide drift-correcting constraints analogous to loop closures. A naive approach (adding every gate observation as a new edge in a pose graph) rapidly inflates the graph, increasing optimization time, and undermining real-time performance.

In this work, we propose a dual pose-graph architecture that addresses this tension. A lightweight \emph{temporary graph} accumulates gate detections at high frequency between main-graph keyframes, capturing the full richness of repeated observations. When a new keyframe is triggered, the temporary graph is optimized, and each gate's multiple observations are distilled into a single refined constraint. Only these compressed constraints are promoted to the \emph{main graph}, which maintains a compact, long-lived representation suitable for incremental optimization and loop closure. The main contributions of this work are:

\begin{enumerate}
    \item A multi-modal, sensor-agnostic semantic SLAM framework that accepts heterogeneous odometry and detection inputs.
    \item A dual pose-graph method that compresses multiple detections into a single optimized edge, preventing graph growth while preserving detection information.
    \item Real flights in a drone racing competition scenario with the system running onboard and online for localization.
\end{enumerate}

\section{Related Work}

\subsection{Visual SLAM and VIO for Drones}

The location of aerial platforms can be broadly divided into three families based on the way visual and inertial data are processed.

\paragraph{Feature-based optimization methods} extract and track sparse key points in frames. ORB-SLAM2~\cite{murtal2017orbslam2} and ORB-SLAM3~\cite{campos2021orbslam3} represent this line of work, combining feature matching with bundle adjustment and loop closure for globally consistent maps. VINS-Mono~\cite{qin2018vins} and VINS-Fusion~\cite{qin2020vinsfusion} extend this paradigm by tightly coupling visual features with IMU pre-integration in a sliding-window optimizer. These systems achieve high accuracy in moderate-speed scenarios but depend on stable feature tracks that degrade under the motion blur and rapid rotations typical of drone racing.

\paragraph{Filter-based methods} propagate state estimates through an EKF, trading global optimality for lower latency. ROVIO~\cite{bloesch2017rovio} matches image patches directly in the filter update, avoiding feature extraction. OpenVINS~\cite{geneva2020openvins} implements a modular MSCKF framework with online camera-IMU calibration, while LARVIO~\cite{qiu2020larvio} targets lightweight deployment on resource-constrained platforms. Filter-based methods are generally faster, but lack the loop closure capability needed to bound long-term drift.

\paragraph{Direct and semi-direct methods} operate on raw image intensities rather than extracted features. SVO~\cite{forster2014svo} combines direct alignment for speed with sparse features for robustness. DM-VIO~\cite{stumberg2022dmvio} extends direct visual odometry with delayed marginalization of the IMU. These methods are sensitive to photometric changes caused by varying lighting conditions on racing tracks.


\subsection{Semantic SLAM}

Semantic SLAM systems augment geometric maps with object-level information, using recognized landmarks as high-level constraints. SLAM++~\cite{salas2013slampp} pioneered this direction by registering known 3D object models as graph nodes. S-Graphs~\cite{bavle2022sgraphs} and its extensions organize structural elements (walls, rooms, floors) into hierarchical scene graphs, while Hydra~\cite{hughes2022hydra} builds multi-layered 3D scene graphs in real time. These approaches demonstrate that semantic landmarks improve robustness and enable higher-level reasoning. However, they target general indoor or outdoor environments and assume slow and exploratory motion. The structured high-speed setting of drone racing, where a small set of distinctive landmarks (gates) repeats along a known track, remains unexplored in this context.

\subsection{Autonomous Drone Racing localization}

Autonomous drone racing has advanced rapidly, with systems that match or exceed human pilots~\cite{kaufmann2023champion}. The AlphaPilot challenge~\cite{foehn2022alphapilot} demonstrated end-to-end autonomous racing combining gate detection with trajectory planning, while the \emph{Race Against the Machine} dataset~\cite{bosello2024race} provides a public benchmark with fully annotated high-speed flights. Gate detection approaches typically employ deep learning models~\cite{li2020gatenet} to localize gate corners in the image, with PnP-based pose estimation providing relative gate positions. However, most racing systems treat detection and localization as decoupled modules: the VIO pipeline estimates the drone state independently, and gate detections are consumed downstream by the planner without feeding back into the state estimate. Our work bridges this gap by incorporating gate detections as first-class landmarks in a pose-graph formulation, tightly coupling them with odometry for drift-corrected localization.

\section{Methodology}


\subsection{Overview}

The proposed method takes as input (i)~odometry estimates from any source (visual odometry, VIO, or other) providing relative pose constraints and (ii)~detections of semantic objects, providing bearing and range measurements to landmarks. The central component is the \emph{Optimizer}, which orchestrates all graph operations: it constructs and manages both the temporary and main graphs, triggers optimization, and provides the estimated drone localization. The method follows a graph-based localization formulation and is designed to be sensor-agnostic; the graph structure is independent of the specific sensors producing the odometry and detection inputs. In the drone racing application, the system operates in a localization mode, where known gate positions are available and incorporated as prior constraints on the landmark nodes.


\subsection{Graph Representation}

The method is formulated as a factor graph composed of two types of nodes and two types of edges:

\begin{itemize}
    \item \textbf{Pose nodes} $\mathbf{x}_t \in SE(3)$: represent the pose of the drone at time~$t$.
    
    \item \textbf{Landmark nodes}: represent semantic objects (e.g., racing gates), which can be modeled as full poses $\mathbf{l}_j \in SE(3)$ or as position-only points $\mathbf{p}_j \in \mathbb{R}^3$.
    
    \item \textbf{Odometry edges}: encode relative pose constraints $\Delta \mathbf{x}_{t,t+1} \in SE(3)$ between consecutive pose nodes, obtained from the odometry source.
    
    \item \textbf{Detection edges}: encode observations that relate a pose node $\mathbf{x}_t$ to a landmark. These can be either (i) full relative pose measurements in $SE(3)$ when the landmark is modeled as a pose, or (ii) position-only constraints in $\mathbb{R}^3$ when the landmark is modeled as a point.
\end{itemize}

The framework allows incorporating prior knowledge of the environment by defining selected landmarks as fixed objects with known positions (e.g., from a pre-existing map). These landmarks are introduced as nodes with prior constraints in the graph. This approach also natively handles multiple simultaneous detections within a single frame.

The optimal trajectory and landmark estimates are obtained by minimizing the sum of squared Mahalanobis distance errors over all edges:
\begin{equation}
    \mathbf{x}^* = \arg\min_{\mathbf{x}, \mathbf{l}} \sum_{(i,j) \in \mathcal{E}_o} \left\| \mathbf{e}_{ij}^{o} \right\|_{\boldsymbol{\Sigma}_{ij}^{o}}^{2} + \sum_{(i,k) \in \mathcal{E}_d} \left\| \mathbf{e}_{ik}^{d} \right\|_{\boldsymbol{\Sigma}_{ik}^{d}}^{2}
    \label{eq:optimization}
\end{equation}
where $\mathcal{E}_o$ and $\mathcal{E}_d$ denote the sets of odometry and detection edges, respectively, $\mathbf{e}$ are the error functions, and $\boldsymbol{\Sigma}$ are the associated covariance matrices encoding measurement uncertainty.

\subsection{Dual Graph System}

\begin{figure}[tb!]
    \centering
    \includegraphics[width=0.95\columnwidth]{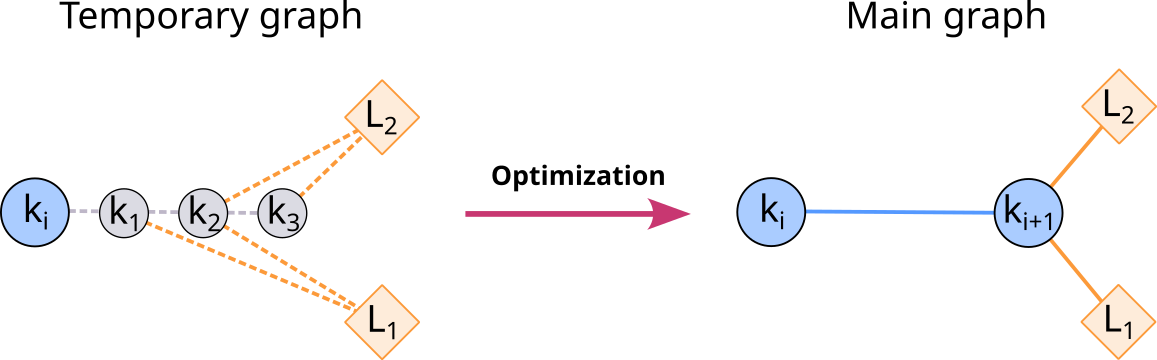}
    \caption{Dual pose-graph architecture. Between main graph keyframes, a temporary graph accumulates detection edges per landmark from high-frequency pose nodes. After optimization, only one refined edge per landmark is promoted to the main graph, keeping it compact while preserving detection information.}
    \label{fig:dual_graph}
    \vspace{-5pt}
\end{figure}

The key contribution of our approach consists of the separation of the pose graph into two complementary structures, illustrated in Fig.~\ref{fig:dual_graph}: the temporary graph and the main graph.

\subsubsection{\textbf{Temporary Graph}}

Between consecutive keyframes of the main graph, a temporary graph is instantiated. This graph operates at a higher frequency, creating pose nodes whenever gate detections are available. Each observation generates a detection edge between the current pose node and the corresponding landmark node, so that a gate observed $N$ times produces $N$ detection edges.

When a new main-graph keyframe is triggered, the temporary graph is optimized. The optimized landmark estimates are then used to construct a single refined detection edge and node per detected landmark, whose information matrix reflects the accumulated confidence from all observations. Those edges are then connected to the new main graph keyframe, and the temporary graph is subsequently discarded, preventing unbounded graph growth.



\subsubsection{\textbf{Main Graph}}

The main graph is the long-lived structure that spans the entire mission. Keyframe pose nodes are added at a controlled rate, and only the distilled single-edge-per-landmark constraints from the temporary graph are integrated. This keeps the main graph compact: instead of accumulating $N$ edges per landmark per keyframe interval, only one high-quality edge is added. The main graph is re-optimized each time a new keyframe is added.

Crucially, the main graph naturally supports loop closure: when a previously existing landmark is detected from a new keyframe, the resulting distilled detection edge connects the current pose to the existing landmark node, providing an implicit loop closure constraint without requiring explicit loop detection.





\section{Implementation}

The system is implemented as a set of ROS~2 nodes within the Aerostack2 \cite{fernandez2023aerostack2} framework, allowing modular integration with existing perception and control stacks. The dual pose graph algorithm has been designed as a ROS 2 node that subscribes to odometry and detection topics and publishes the optimized pose and the transformation between the \textit{map} and the \textit{odom} frames. Graph optimization is performed using the \texttt{g2o} library~\cite{kummerle2011g2o} with a Levenberg--Marquardt optimizer and CHOLMOD-based sparse Cholesky factorization. 
The temporary graph accumulates detections between consecutive keyframes of the main graph, and is optimized when a new keyframe is added; its refined estimates are then incorporated into the main graph, after which it is reset, and the main graph is re-optimized. In practice, optimization is limited to a fixed number of iterations to ensure real-time performance. Keyframes are created when the traveled distance exceeds the predefined threshold $d_{\text{main}}$ and $d_{\text{temp}}$ for each graph.


\subsection{Gate Detection and Data Association}

Gate detection follows the approach of AlphaPilot~\cite{foehn2022alphapilot}. An input image is processed by a neural network that predicts corner confidence maps and path affinity fields, enabling robust multi-gate detection. The relative pose of each gate is then estimated using Perspective-n-Point (PnP) with known gate dimensions, and the pipeline is optimized with TensorRT for efficient onboard execution.

Detected gates are transformed to the global frame and filtered to remove geometrically inconsistent observations. Data association is performed using the Hungarian algorithm based on Euclidean distance, with outliers rejected via thresholding. Additional orientation checks ensure consistency with expected gate directions, including handling reverse observations. Accepted detections are assigned unique identifiers and used to update the corresponding landmarks.

\subsection{Visual-Inertial Odometry}

Visual-inertial odometry is provided by OpenVINS~\cite{geneva2020openvins}, 
an open-source EKF-based visual-inertial estimator that fuses monocular camera and IMU measurements using a Multi-State Constraint Kalman Filter (MSCKF) formulation. OpenVINS provides robust state estimation under aggressive motion through online calibration of camera-IMU extrinsics and time offsets, making it well-suited for the high-speed dynamics of drone racing.

\section{Experimental Validation}

\subsection{Experimental Setup}

To validate the dual graph design, we perform an \textbf{ablation study} comparing it against a single-graph variant that adds all gate detection edges directly to the main graph without the temporary graph compression step. We evaluated two odometry threshold configurations to assess scalability.

We evaluated the proposed system on the \emph{Race Against the Machine} \textbf{drone racing dataset}~\cite{bosello2024race}, which provides synchronized high-resolution images with gate annotations, IMU data at 500 \, Hz, and motion-capture ground-truth poses, following the protocol of~\cite{azhari2025drift}, with a downsampled image stream - dropping 3 out of every 4 images - and camera calibration using the built-in online calibration system in OpenVINS. We use the set of piloted sequences \texttt{ellipse} and \texttt{lemniscate}, 3 different runs for each one.

Finally, we validated the system on \textbf{real-world flight} sequences collected during the A2RL drone racing competition in April 2025, each consisting of two laps through 11 gates on a \SI{170}{\meter} track at high speeds. In these flights, the system ran onboard a drone equipped with an NVIDIA\regsym~Jetson Orin NX, an IMU, and a monocular rolling-shutter camera.

\subsubsection{\textbf{Metrics}}

For the ablation study, we report Absolute Trajectory Error (ATE) and the graph optimization time to evaluate the trade-off between accuracy and computational cost.
For the drone racing dataset, we report ATE as the root-mean-square error after $SE(3)$ alignment.
For real-world experiments, where ground truth is unavailable, we evaluated performance using the correction magnitude between the graph-based estimate and VIO at gate crossings, computed as the Euclidean distance to the closest trajectory point to each gate. Results are aggregated per lap and across sequences.



\begin{table}[b]
\centering
\caption{Ablation study of dual-graph and single-graph architectures on    
Flight 01p Ellipse sequence.}
\setlength{\tabcolsep}{3pt}
\begin{tabular}{l S[table-format=1.1] S[table-format=1.1] S[table-format=1.3]
S[table-format=4.0] S[table-format=4.0] S[table-format=3.1]}
\toprule
{Graph} & {$d_{\text{main}}$ (m)} & {$d_{\text{temp}}$ (m)} & {ATE (m)} & {Nodes} & {Edges} & {Opt P95 (ms)} \\
\midrule
Dual   & 2.0 & 0.5  & 0.563          & 250  & 1213 & 57.7  \\
Dual   & 2.0 & 0.1  & 0.548          & 249  & 1188 & 50.1  \\
Dual   & 0.5 & 0.1  & 0.502          & 992  & 4518 & 170.9 \\
Single & 2.0 & {--} & 0.635          & 248  & 1203 & 62.7  \\
Single & 0.5 & {--} & 0.537          & 991  & 4890 & 186.7 \\
Single & 0.1 & {--} & \bfseries 0.346 & 1822 & 8617 & 219.9 \\               
\bottomrule
\end{tabular}
\label{tab:ellipse_jetson_results}
\end{table} 



\begin{table}[b]
\centering

\caption{ATE (mean $\pm$ std) on racing dataset sequences.}

\begin{minipage}{\linewidth}
\centering
\textbf{Ellipse sequence}
\setlength{\tabcolsep}{10pt}
\begin{tabular}{lcc}
\toprule
Method & ATE Trans (\si{\meter}) & ATE Rot (\si{\degree}) \\
\midrule
OpenVINS & $1.075 \pm 0.465$ & $5.51 \pm 3.90$ \\
Ours (\SI{2.0}{\meter}/\SI{0.1}{\meter}) & $0.463 \pm 0.073$ & $5.32 \pm 0.81$ \\
Ours (\SI{0.5}{\meter}/\SI{0.1}{\meter}) & $\mathbf{0.412 \pm 0.086}$ & $\mathbf{4.41 \pm 0.71}$ \\
\bottomrule
\end{tabular}
\end{minipage}

\vspace{1.0em}

\begin{minipage}{\linewidth}
\centering
\textbf{Lemniscate sequence}
\setlength{\tabcolsep}{10pt}
\begin{tabular}{lcc}
\toprule
Method & ATE Trans (\si{\meter}) & ATE Rot (\si{\degree}) \\
\midrule
OpenVINS & $1.367 \pm 0.712$ & $10.69 \pm 8.24$ \\
Ours (\SI{2.0}{\meter}/\SI{0.1}{\meter}) & $0.381 \pm 0.110$ & $9.87 \pm 7.66$ \\
Ours (\SI{0.5}{\meter}/\SI{0.1}{\meter}) & $\mathbf{0.327 \pm 0.084}$ & $\mathbf{9.87 \pm 8.76}$ \\
\bottomrule
\end{tabular}
\end{minipage}

\label{tab:dataset_ate}
\end{table}

\begin{figure}[htbp]           
  \centering
  \begin{subfigure}[b]{\columnwidth}
  \centering
  \includegraphics[width=\textwidth]{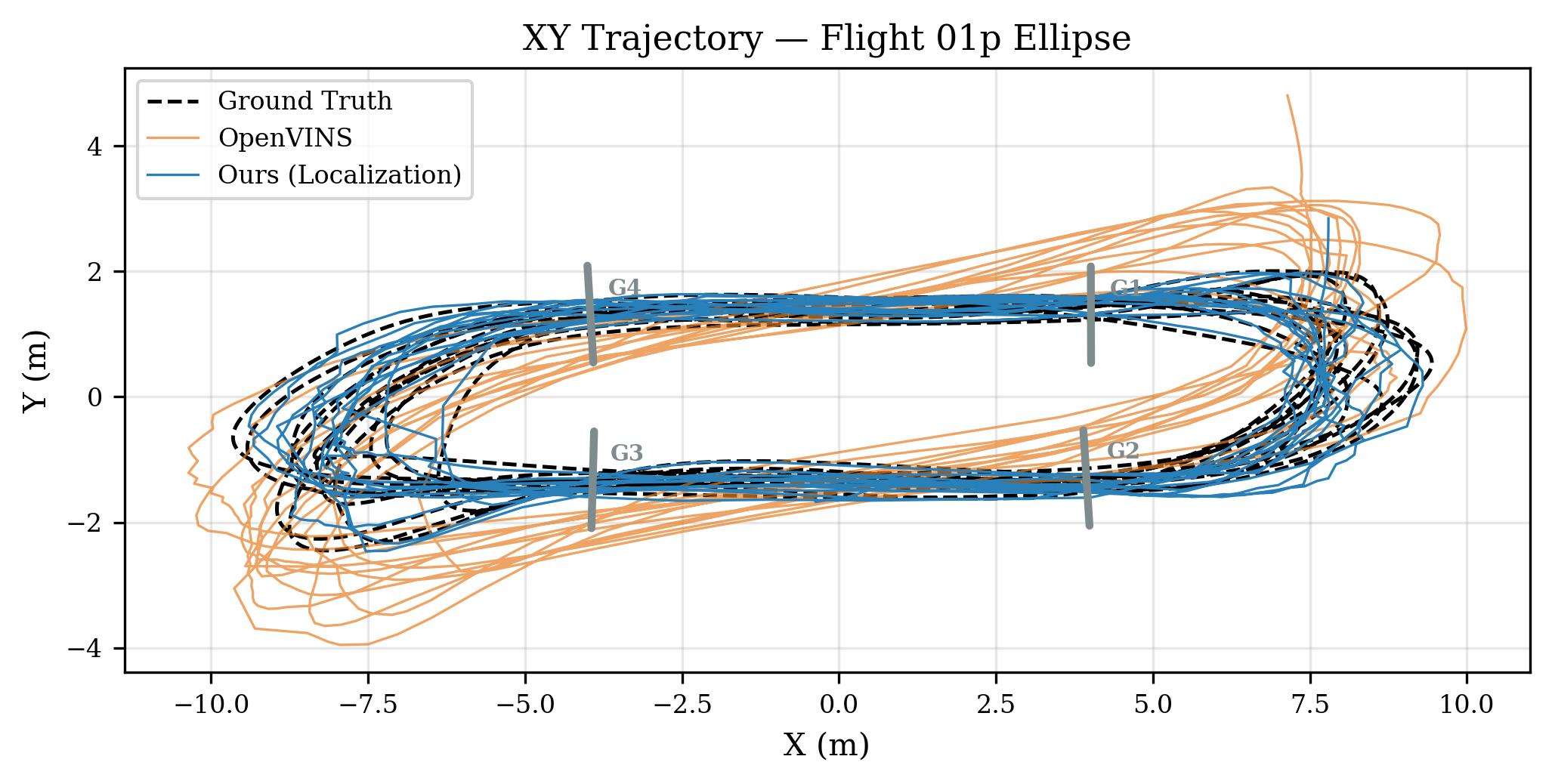}           
  \label{fig:traj_ellipse}
  \end{subfigure}
  \vspace{-8pt}
  \begin{subfigure}[b]{\columnwidth} 
      \centering
      \includegraphics[width=\textwidth]{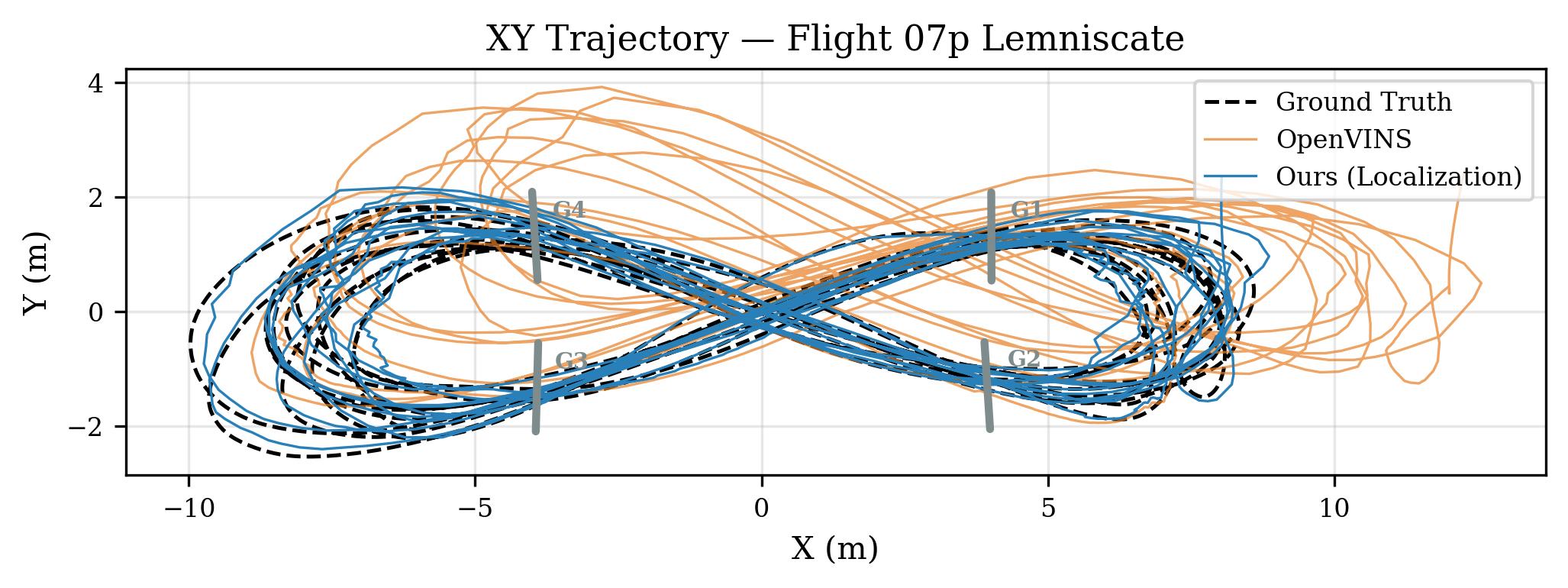}
      \label{fig:traj_lemniscate}
  \end{subfigure}
  \vspace{-8pt}
  \caption{XY trajectory comparison on ellipse and lemniscate tracks.
Ground truth (dashed black), OpenVINS VIO (orange), and our corrected        
localization (blue). Gray rectangles indicate the racing gates used as     
landmarks.}
  \vspace{-8pt}
\label{fig:trajectories}
\end{figure}

\subsection{Results and discussion}





\subsubsection{\textbf{Ablation Study}} As shown in Table~\ref{tab:ellipse_jetson_results}, the dual-graph architecture improves accuracy without increasing the size of the graph or the computational cost by leveraging the information accumulated in the temporary graph. For comparable graph sizes, the dual graph consistently achieves lower ATE than the single-graph formulation.


Furthermore, the dual-graph formulation benefits from increased information accumulation: reducing the temporary graph threshold (from \SI{0.5}{\meter} to \SI{0.1}{\meter}) leads to improved accuracy (\SI{0.563}{\meter} to \SI{0.548}{\meter}) without increasing the size of the main graph. At comparable keyframe frequencies ($d_{\text{th}}=$ \SI{2}{\meter} and \SI{0.5}{\meter}), the dual graph consistently outperforms the single-graph approach, achieving lower ATE while maintaining smaller graphs and shorter optimization time.

These results demonstrate that aggregating observations in the temporary graph produces more informative and consistent constraints in the main       
graph, improving the accuracy of the estimation without requiring denser graphs. In contrast, the single-graph approach must increase graph density   
(up to 1822 nodes and 8617 edges) and computational cost (\SI{219.9}{\milli\second}) to achieve its best accuracy.                    
           
A limitation of the dual-graph formulation is that optimization is           
performed less frequently due to keyframe-based integration, which may     
reduce correction responsiveness compared to the single-graph approach       
under very high update rates.

\subsubsection{\textbf{Drone racing dataset}}

Table~\ref{tab:dataset_ate} reports the ATE results on the ellipse and lemniscate sequences. Our approach is evaluated in two configurations: $d_{\text{main}} =$ \SI{2.0}{\meter} and $d_{\text{main}} =$ \SI{0.5}{\meter}, both with $d_{\text{temp}} =$ \SI{0.1}{\meter}.
The proposed method consistently improves over the OpenVINS baseline, reducing the ATE by more than 50\% in both scenarios. For the ellipse sequence, the ATE decreases from \SI{1.075}{\meter} to \SI{0.412}{\meter}, while for the lemniscate sequence it is reduced from \SI{1.367}{\meter} to \SI{0.327}{\meter}. These results demonstrate the effectiveness of the approach in mitigating the drift in the odometry, illustrated in Fig.~\ref{fig:trajectories}.

\begin{figure}[t]
    \centering
    \includegraphics[width=0.95\columnwidth]{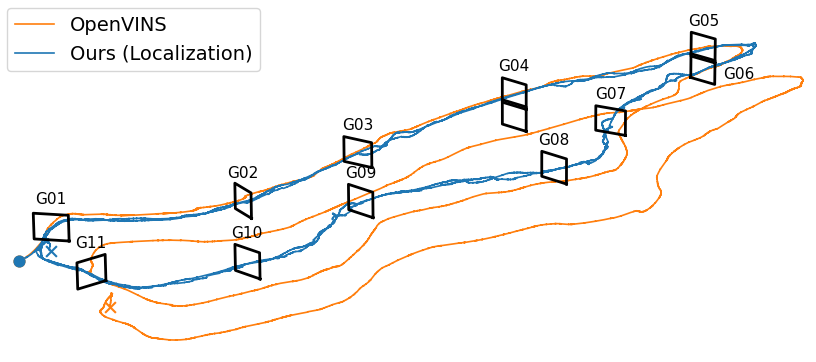}
    \caption{Comparison between OpenVINS odometry and our corrected localization during a real flight in A2RL competition with the system running onboard.}
    \label{fig:a2rl_trajectory}
\end{figure}




\begin{table}[t]
\centering
\caption{Correction magnitude between OpenVINS baseline and our localization at gate crossings (in meters). Values are reported as mean $\pm$    
standard deviation per lap.}
\label{tab:drift_eval}
\begin{tabular}{lccc}
\toprule
Race Flight & Lap 1 (\si{\meter}) & Lap 2 (\si{\meter}) & Full Track (\si{\meter}) \\
\midrule
Flight 1 & $1.570 \pm 0.946$ & $5.375 \pm 1.824$ & $3.472 \pm 1.385$ \\
Flight 2 & $1.393 \pm 0.835$ & $4.527 \pm 1.839$ & $2.960 \pm 1.337$ \\      
Flight 3 & $1.512 \pm 0.649$ & $2.778 \pm 1.061$ & $2.145 \pm 0.855$ \\    
\midrule
Average  & $\mathbf{1.492 \pm 0.073}$ & $\mathbf{4.227 \pm 1.081}$ &       
$\mathbf{2.859 \pm 0.577}$ \\
\bottomrule
\end{tabular}
\end{table} 

\subsubsection{\textbf{Real-world flights}} Table \ref{tab:drift_eval} shows the results obtained during the competition. The correction magnitude increases over laps, indicating accumulated drift in the raw odometry. On average, it grows from about \SI{1.5}{\meter} in the first lap to more than \SI{4}{\meter} in the second. This shows that the proposed system effectively compensates for drift and maintains global consistency over long trajectories. The relatively low variance across sequences suggests stable performance under varying conditions. Despite not measuring absolute accuracy, this metric provides a practical proxy for drift and correction effectiveness in real-flight scenarios. Fig. \ref{fig:a2rl_trajectory} shows a visual comparison between the two estimates during one of the flights.

\section{Conclusion}


We presented a dual pose-graph SLAM architecture for vision-based localization
in autonomous drone racing. By routing gate detections through a temporary 
graph before merging refined constraints into the main graph, the system
achieves \SIrange{10}{12}{\percent} better translational accuracy than a
single-graph baseline at identical computational cost. 
Experimental validation on the drone-racing dataset shows an ATE reduction of \SIrange{56}{74}{\percent} over standalone VIO.
The deployment in the A2RL competition demonstrated that the system operates fully onboard, providing online localization during real flights, reducing the drift of the odometry baseline by up to 4.2 meters.
Future work will validate the sensor-agnostic capabilities of the framework with additional odometry and detection sources, and other robotics applications.


\section*{Acknowledgment}

This work was supported by Horizon Europe projects SHEREC from the European Union "Safe Healthy and Environmental Ship Recycling", ref: 101136056, and “CORESENSE: A Hybrid Cognitive Architecture for Deep Understanding”, ref: 101070254. It has also been supported by the project "INSERTION", ref. ID2021-127648OBC32, and the work of the first author is supported by the Program for Technical Assistants PTA2021-020671, both funded by the Spanish Ministry of Science and Innovation.

\bibliographystyle{IEEEtran}
\bibliography{bibliography}

@article{kaufmann2023champion,
  title={{Champion-level drone racing using deep reinforcement learning}},
  author={Kaufmann, Thomas and Bauersfeld, Leonard and Loquercio, Antonio and M{\"u}ller, Matthias and Koltun, Vladlen and Scaramuzza, Davide},
  journal={Nature},
  volume={620},
  pages={982--987},
  year={2023}
}

@article{foehn2022alphapilot,
  title={{Alphapilot: Autonomous drone racing}},
  author={Foehn, Philipp and Brescianini, Dario and Kaufmann, Elia and Cieslewski, Titus and Gehrig, Mathias and Muglikar, Mayur and Scaramuzza, Davide},
  journal={Auton. Robots},
  volume={46},
  number={1},
  pages={307--320},
  year={2022}
}

@article{campos2021orbslam3,
  title={{ORB-SLAM3}: An accurate open-source library for visual, visual-inertial, and multimap SLAM},
  author={Campos, Carlos and Elvira, Richard and Rodr{\'i}guez G{\'o}mez, Juan J. and Montiel, Jose M. M. and Tard{\'o}s, Juan D.},
  journal={IEEE Trans. Robot.},
  volume={37},
  number={6},
  pages={1874--1890},
  year={2021}
}

@article{qin2018vins,
  title={{VINS-Mono}: A robust and versatile monocular visual-inertial state estimator},
  author={Qin, Tong and Li, Peiliang and Shen, Shaojie},
  journal={IEEE Trans. Robot.},
  volume={34},
  number={4},
  pages={1004--1020},
  year={2018}
}

@article{qin2020vinsfusion,
  title={{A general optimization-based framework for local odometry estimation with multiple sensors}},
  author={Qin, Tong and Pan, Jiarong and Cao, Shaozu and Shen, Shaojie},
  journal={Auton. Robots},
  volume={44},
  number={3},
  pages={421--436},
  year={2020}
}

@inproceedings{forster2014svo,
  title={{SVO}: Fast semi-direct monocular visual odometry},
  author={Forster, Christian and Pizzoli, Matia and Scaramuzza, Davide},
  booktitle={Proc. IEEE Int. Conf. Robot. Autom. (ICRA)},
  pages={15--22},
  year={2014}
}

@article{murtal2017orbslam2,
  title={{ORB-SLAM2}: An open-source SLAM system for monocular, stereo, and RGB-D cameras},
  author={Mur-Artal, Raul and Tard{\'o}s, Juan D.},
  journal={IEEE Trans. Robot.},
  volume={33},
  number={5},
  pages={1255--1262},
  year={2017}
}

@inproceedings{salas2013slampp,
  title={{SLAM++}: Simultaneous localisation and mapping at the level of objects},
  author={Salas-Moreno, Renato F. and Newcombe, Richard A. and Strasdat, Hauke and Kelly, Paul H. J. and Davison, Andrew J.},
  booktitle={Proc. IEEE Conf. Comput. Vis. Pattern Recognit. (CVPR)},
  pages={1352--1359},
  year={2013}
}

@article{bavle2022sgraphs,
  title={{Situational graphs for robot navigation in structured indoor environments}},
  author={Bavle, Hriday and Sanchez-Lopez, Jose Luis and Shaheer, Mohammad and Civera, Javier and Voos, Holger},
  journal={IEEE Robot. Autom. Lett.},
  volume={7},
  number={4},
  pages={9107--9114},
  year={2022}
}

@inproceedings{hughes2022hydra,
  title={{Hydra}: A real-time spatial perception system for 3D scene graph construction and optimization},
  author={Hughes, Nathan and Chang, Yun and Carlone, Luca},
  booktitle={Robotics: Science and Systems (RSS)},
  year={2022}
}

@inproceedings{li2020gatenet,
  title={{Autonomous drone race: A computationally efficient gate detection and path planning}},
  author={Li, Shushuai and De Wagter, Christophe and de Croon, Guido C. H. E.},
  booktitle={Proc. IEEE/RSJ Int. Conf. Intell. Robots Syst. (IROS)},
  pages={3330--3336},
  year={2020}
}

@inproceedings{kummerle2011g2o,
  title={{g2o}: A general framework for graph optimization},
  author={K{\"u}mmerle, Rainer and Grisetti, Giorgio and Strasdat, Hauke and Konolige, Kurt and Burgard, Wolfram},
  booktitle={Proc. IEEE Int. Conf. Robot. Autom. (ICRA)},
  pages={3607--3613},
  year={2011}
}

@inproceedings{geneva2020openvins,
  title={{OpenVINS}: A research platform for visual-inertial estimation},
  author={Geneva, Patrick and Eckenhoff, Kevin and Lee, Woosik and Yang, Yulin and Huang, Guoquan},
  booktitle={Proc. IEEE Int. Conf. Robot. Autom. (ICRA)},
  pages={4328--4334},
  year={2020}
}

@article{bosello2024race,
  title={{Race against the machine: A fully-annotated, open-design dataset of autonomous and piloted high-speed flight}},
  author={Bosello, Michele and Aguiari, Daniele and Bertogna, Massimo and Mottola, Luca},
  journal={IEEE Robot. Autom. Lett.},
  volume={9},
  number={4},
  pages={3799--3806},
  year={2024}
}

@article{bloesch2017rovio,
  title={{Iterated extended Kalman filter based visual-inertial odometry using direct photometric feedback}},
  author={Bloesch, Michael and Burri, Michael and Omari, Sammy and Hutter, Marco and Siegwart, Roland},
  journal={Int. J. Robot. Res.},
  volume={36},
  number={10},
  pages={1053--1072},
  year={2017}
}

@article{qiu2020larvio,
  title={{Tracking at least 3 keypoints in a single image with lightweight, accurate and robust monocular visual-inertial SLAM}},
  author={Qiu, Ke and Qin, Tong and Pan, Jiarong and Liu, Shuaijun and Shen, Shaojie},
  journal={IEEE Robot. Autom. Lett.},
  volume={5},
  number={4},
  pages={5257--5264},
  year={2020}
}

@article{stumberg2022dmvio,
  title={{DM-VIO}: Delayed marginalization visual-inertial odometry},
  author={von Stumberg, Lukas and Cremers, Daniel},
  journal={IEEE Robot. Autom. Lett.},
  volume={7},
  number={2},
  pages={1408--1415},
  year={2022}
}

@article{azhari2025drift,
  title={{Drift-corrected monocular VIO and perception-aware planning for autonomous drone racing}},
  author={Azhari, A. and Bosello, M. and Aguiari, D. and Mottola, L.},
  journal={arXiv},
  eprint={2512.20475},
  year={2025}
}

@article{fernandez2023aerostack2,
  title={{Aerostack2}: A software framework for developing multi-robot aerial systems},
  author={Fernandez-Cortizas, Miguel and Molina, Martin and Arias-Perez, Pedro and Perez-Segui, Rafael and Perez-Saura, David and Campoy, Pascual},
  journal={arXiv},
  eprint={2303.18237},
  year={2023}
}

\end{document}